\documentclass{article}

\pdfoutput=1

 \PassOptionsToPackage{square,numbers}{natbib}
 
 \usepackage[final]{nips_2016}

\usepackage{amsfonts}       
\usepackage{amsmath}	
\DeclareMathOperator{\rec}{rec}

\usepackage{graphicx}
\usepackage{caption}
\usepackage[labelformat=simple]{subcaption}
\usepackage{multirow}
\usepackage{adjustbox}

\usepackage{etoolbox,siunitx}
\robustify\bfseries   					
\usepackage[symbol]{footmisc}

\usepackage{authblk} 
 
\usepackage{hyperref}       

\usepackage[symbol]{footmisc}

\title{\Large{DeepSurv: Personalized Treatment Recommender System Using A Cox Proportional Hazards Deep Neural Network}}

\author[1]{Jared L. Katzman}
\author[2,5]{Uri Shaham}
\author[3,8]{Alexander Cloninger}
\author[3,4,5]{Jonathan Bates}
\author[6]{Tingting Jiang}
\author[3,6,7]{Yuval Kluger}
\affil[1]{Department of Computer Science, Yale University, 51 Prospect Steet, New Haven, CT 06511, USA}
\affil[2]{Department of Statistics, Yale University, 24 Hillhouse Avenue, New Haven, CT 06511, USA}
\affil[3]{Applied Mathematics Program, Yale University, 51 Prospect Steet, New Haven, CT 06511, USA}
\affil[4]{Yale School of Medicine, 333 Cedar Street, New Haven CT 06510, USA}
\affil[5]{Center for Outcomes Research and Evaluation, Yale-New Haven Hospital, New Haven, CT}
\affil[6]{Interdepartmental Program in Computational Biology and Bioinformatics, Yale University, New Haven, CT 06511, USA}
\affil[7]{Department of Pathology and Yale Cancer Center, Yale University School of Medicine, New Haven, CT, USA}
\affil[8]{Department of Mathematics, University of California, San Diego, La Jolla, CA 92093, USA}

\begin{document}

\maketitle

\begin{abstract} Medical practitioners use survival models to explore and understand the relationships between patients' covariates (e.g. clinical and genetic features) and the effectiveness of various treatment options. Standard survival models like the linear Cox proportional hazards model require extensive feature engineering or prior medical knowledge to model treatment interaction at an individual level. While nonlinear survival methods, such as neural networks and survival forests, can inherently model these high-level interaction terms, they have yet to be shown as effective treatment recommender systems. We introduce DeepSurv, a Cox proportional hazards deep neural network and state-of-the-art survival method for modeling interactions between a patient's covariates and treatment effectiveness in order to provide personalized treatment recommendations. We perform a number of experiments training DeepSurv on simulated and real survival data. We demonstrate that DeepSurv performs as well as or better than other state-of-the-art survival models and validate that DeepSurv successfully models increasingly complex relationships between a patient's covariates and their risk of failure. We then show how DeepSurv models the relationship between a patient's features and effectiveness of different treatment options to show how DeepSurv can be used to provide individual treatment recommendations. Finally, we train DeepSurv on real clinical studies to demonstrate how it's personalized treatment recommendations would increase the survival time of a set of patients. The predictive and modeling capabilities of DeepSurv will enable medical researchers to use deep neural networks as a tool in their exploration, understanding, and prediction of the effects of a patient's characteristics on their risk of failure. \\
\end{abstract}

\section{Introduction}
Medical researchers use survival models to evaluate the significance of prognostic variables in outcomes such as death or cancer recurrence and subsequently inform patients of their treatment options \citep{YehJAMA, royston2013external, bair2004semi,cheng2013development}. One standard survival model is the Cox proportional hazards model (CPH)  \citep{cox1992regression}. The CPH is a semiparametric model that calculates the effects of observed covariates on the risk of an event occurring (e.g. `death'). The model assumes that a patient's risk of failure is a linear combination of the patient's covariates. This assumption is referred to as the \textit{linear proportional hazards} condition. However, in many applications, such as providing personalized treatment recommendations, it may be too simplistic to assume that the risk function is linear. As such, a richer family of survival models is needed to better fit survival data with nonlinear risk functions. 

To model nonlinear survival data, researchers have applied three main types of neural networks to the problem of survival analysis. These include variants of: (i) classification methods \citep[see details in][]{liestbl1994survival,street1998neural}, (ii) time-encoded methods \citep[see details in][]{franco2005artificial, biganzoli1998feed}, (iii) and risk-predicting methods \citep[see details in][]{faraggi1995neural}. This third type is a feed-forward neural network (NN) that estimates an individual's risk of failure. In fact, Faraggi-Simon's network is seen as a nonlinear extension of the Cox proportional hazards model.

Risk neural networks learn highly complex and nonlinear relationships between prognostic features and an individual's risk of failure. In application, for example, when the success of a treatment option is affected by an individual's features, the NN learns the relationship without prior feature selection or domain expertise. The network is then able to provide a personalized recommendation based on the computed risk of a treatment.

However, previous studies have demonstrated mixed results on NNs ability to predict risk. For instance, researchers have attempted to apply the Faraggi-Simon network with various extensions, but they have failed to demonstrate improvements beyond the linear Cox model, see \cite{sargent2001comparison}, \cite{xiang2000comparison} and \cite{mariani1997prognostic}. One possible explanation is that the practice of NNs was not as developed as it is today. To the best of our knowledge, NNs have not outperformed standard methods for survival analysis (e.g. CPH). Our manuscript shows that this is no longer the case; with modern techniques, risk NNs have state-of-the-art performance and can be used for a variety of medical applications. 

The goals of this paper are: (i) to show that the application of deep learning to survival analysis performs as well as or better than other survival methods in predicting risk; and (ii) to demonstrate that the deep neural network can be used as a personalized treatment recommender system and a useful framework for further medical research. 

We propose a modern Cox proportional hazards deep neural network, henceforth referred to as DeepSurv, as the basis for a treatment recommender system. We make the following contributions. First, we show that DeepSurv performs as well as or better than other survival analysis methods on survival data with both linear and nonlinear risk functions. Second, we include an additional categorical variable representing a patient's treatment group to illustrate how the network can learn complex relationships between an individual's covariates and the effect of a treatment. Our experiments validate that the network successfully models the treatment's risk within a population. Third, we use DeepSurv to provide treatment recommendations tailored to a patient's observed features. We confirm our results on real clinical studies, which further demonstrates the power of DeepSurv. Finally, we show that the recommender system supports medical practitioners in providing personalized treatment recommendations that potentially could increase the median survival time for a set of patients.

The organization of the manuscript is as follows: in Section \ref{sec:background}, we provide a brief background on survival analysis. In Section \ref{sec:dcph}, we present our contributions, including an explanation of our implementation of DeepSurv and our proposed recommender system. In Section \ref{sec:experiments}, we describe the experimental design and results. Section \ref{sec:conclusion} concludes the manuscript. 

\section{Background} \label{sec:background}

In this section, we define survival data and the approaches for modeling a population's survival and failure rate. Additionally, we discuss linear and nonlinear survival models and their limitations.

\subsection{Survival data} 
Survival data is comprised of three elements: a patient's baseline data $x$, a failure event time $T$, and an event indicator $E$. If an event (e.g. death) is observed, the time interval $T$ corresponds to the time elapsed between the time in which the baseline data was collected and the time of the event occurring, and the event indicator is $E=1$. If an event is not observed, the time interval $T$ corresponds to the time elapsed between the collection of the baseline data and the last contact with the patient (e.g. end of study), and the event indicator is $E=0$. In this case, the patient is said to be \textit{right-censored}. If one opts to use standard regression methods, the right-censored data is considered to be a type of missing data. This is typically discarded which can introduce a bias in the model. Therefore, modeling right-censored data requires special consideration or the use of a survival model. 

Survival and hazard functions are the two fundamental functions in survival analysis. The survival function is denoted by $S(t) = \Pr(T > t)$, which signifies the probability that an individual has `survived' beyond time $t$. The hazard function is a measure of risk at time $t$. A greater hazard signifies a greater risk of death. The hazard function $\lambda(t)$ is defined as:
\begin{equation}
\lambda(t) = \underset{\delta \rightarrow 0}{\lim} \: \frac{\Pr(t \leq T < t + \delta \: | \: T \geq t)}{\delta}.
\end{equation}

A proportional hazards model is a common method for modeling an individual's survival given their baseline data $x$. The model assumes that the hazard function is composed of two functions: a baseline hazard function, $\lambda_0(t)$, and a risk function, $h(x)$, denoting the effects of an individual's covariates. The hazard function is assumed to have the form 
$\lambda(t | x) = \lambda_0(t) \cdot e ^ {h(x)}$. 

\subsection{Linear Survival Models} 

The CPH is a proportional hazards model that estimates the risk function $h(x)$ by a linear function $\hat{h}_\beta(x) = \beta^Tx$. To perform Cox regression, one tunes the weights $\beta$ to optimize the Cox partial likelihood. The partial likelihood is the product of the probability at each event time $T_i$ that the event has occurred to individual $i$, given the set of individuals still at risk at time $T_i$. The Cox partial likelihood is parameterized by $\beta$ and defined as
\begin{equation} \label{eq:cox-likelihood} 
L_c(\beta) = \underset{i : E_i = 1}{\prod} \frac{\exp (\hat{h}_\beta(x_i))} { \underset{j \in \Re(T_i)}{\sum} \exp ( \hat{h}_\beta(x_j))},
\end{equation}
where the values $T_i$, $E_i$, and $x_i$ are the respective event time, event indicator, and baseline data for the $i^{th}$ observation. The product is defined over the set of patients with an observable event $E_i = 1$. The risk set $\Re(t) = \{ i : T_i \geq t \}$ is the set of patients still at risk of failure at time $t$.

In many applications, for example modeling nonlinear gene interactions, we cannot assume the data satisfies the linear proportional hazards condition. In this case, the CPH model would require computing high-level interaction terms. This becomes prohibitively expensive as the number of features and interactions increases. Therefore, a more complex nonlinear model is needed.

\subsection {Nonlinear survival models}

The Faraggi-Simon method is a feed-forward neural network that provides the basis for a nonlinear proportional hazards model. \cite{faraggi1995neural} experimented with a single hidden layer network with two or three nodes. Their model requires no prior assumption of the risk function $h(x)$ other than continuity. Instead, the NN computes nonlinear features from the training data and calculates their linear combination to estimate the risk function. Similar to Cox regression, the network optimizes a modified Cox partial likelihood. They replace the linear combination of features $\hat{h}_\beta(x)$ in Equation \ref{eq:cox-likelihood} with the output of the network $\hat{h}_\theta(x)$.

As previous research suggests, the Faraggi-Simon network has not been shown to outperform the linear CPH \citep{faraggi1995neural,xiang2000comparison,mariani1997prognostic}. Furthermore, to the best of our knowledge, we were the first to attempt applying modern deep learning techniques to the Cox proportional hazards loss function. 

Another popular machine learning approach to modeling patients' risk function is the random survival forest (RSF) \citep{rsf2,rsf3}. The random survival forest is a tree method that produces an ensemble estimate for the cumulative hazard function. 

A more recent deep learning approach models the event time according to a Weibull distribution with parameters given by latent variables generated by a deep exponential family \citep{bleiDSA}.

\section{Methods} \label{sec:dcph}

In this section, we describe our methodology for providing personalized treatment recommendations using DeepSurv. First, we describe the architecture and training details of DeepSurv, an open source Python module that applies recent deep learning techniques to a nonlinear Cox proportional hazards network. Second, we define DeepSurv as a prognostic model and show how to use the network's predicted risk function to provide personalized treatment recommendations. 

\subsection{DeepSurv}
DeepSurv is a multi-layer perceptron, which predicts a patient's risk of death. The output of the network is a single node, which estimates the risk function $\hat{h}_\theta(x)$ parameterized by the weights of the network $\theta$. Similar to the Faraggi-Simon network, we set the loss function to be the negative log partial likelihood of Equation \ref{eq:cox-likelihood}:
 \begin{equation} \label{eq:loss-function}
l(\theta) := -\sum_{i: E_i = 1} \Big( \hat{h}_\theta(x_i) - \log \sum_{j \in \Re(T_i)} e^{\hat{h}_\theta(x_j)} \Big). 
\end{equation}
	
We allow a deep architecture (i.e. more than one hidden layer) and apply modern techniques such as weight decay regularization, Rectified Linear Units (ReLU) \citep{nair2010rectified} with batch normalization \citep{ioffe2015batch}, Scaled Exponential Linear Units (SELU) \citep{gunter2017self}, dropout \citep{srivastava2014dropout}, gradient descent optimization algorithms (Stochastic Gradient Descent and Adaptive Moment Estimation (Adam) \citep{KingmaB14}), Nesterov momentum \citep{nesterov2013gradient}, gradient clipping \citep{pascanu2012understanding}, and learning rate scheduling \citep{senior2013empirical}.

To tune the network's hyper-parameters, we perform a Random hyper-parameter optimization search \citep{bergstra2012random}. For more technical details, see Appendix \ref{sec:experimental-details}. 

\subsection{Treatment recommender system} \label{sec:treatment_rec}



In a clinical study, patients are subject to different levels of risk based on their relevant prognostic features and which treatment they undergo. We generalize this assumption as follows. Let all patients in a given study be assigned to one of $n$ treatment groups $\tau \in \{0, 1, ..., n-1\}$. We assume each treatment $i$ to have an independent risk function $h_i(x)$. Collectively, the hazard function becomes:

\begin{equation}
\lambda(t; x | \tau = i) = \lambda_0(t) \cdot e ^ {h_i(x)}.
	\label{recommender-exp-hr}
\end{equation}

For any patient, the network should be able to accurately predict the risk $h_i(x)$ of being prescribed a given treatment $i$. Then, based on the assumption that each individual has the same baseline hazard function $\lambda_{0}(t)$, we can take the log of the hazards ratio to calculate the personal risk of prescribing one treatment option over another. We define this difference of log hazards as the \textit{recommender} function or $\rec_{ij}(x)$:

\begin{equation} \begin{aligned} \label{eq:rec(x)}
\rec_{ij}(x) &= \log \Big( \frac{\lambda(t;x | \tau = i)} {\lambda(t; x | \tau = j)} \Big) = \log \Big( \frac{\lambda_0(t) \cdot e^{h_i(x)}}{\lambda_0(t) \cdot e^{h_j(x)}} \Big) \\
&= h_i(x) - h_j(x).
\end{aligned}
\end{equation}

The recommender function can be used to provide personalized treatment recommendations. We first pass a patient through the network once in treatment group $i$ and again in treatment group $j$ and take the difference. When a patient receives a positive recommendation $\rec_{ij}(x)$, treatment $i$ leads to a higher risk of death than treatment $j$. Hence, the patient should be prescribed treatment $j$. Conversely, a negative recommendation indicates that treatment $i$ is more effective and leads to a lower risk of death than treatment $j$, and we recommend treatment $i$. 

DeepSurv's architecture holds an advantage over the CPH because it calculates the recommender function without an \textit{a priori} specification of treatment interaction terms. In contrast, the CPH model computes a constant recommender function unless treatment interaction terms are added to the model, see Appendix \ref{appendix:cph} for more details. Discovering relevant interaction terms is expensive because it requires extensive experimentation or prior biological knowledge of treatment outcomes. Therefore, DeepSurv is more cost-effective compared to CPH. 

\section{Results}

\label{sec:experiments}

We perform four sets of experiments: (i) simulated survival data, (ii) real survival data, (iii) simulated treatment data, and (iv) real treatment data. First, we use simulated data to show how DeepSurv successfully learns the true risk function of a population. Second, we validate the network's predictive ability by training DeepSurv on real survival data. Third, we simulate treatment data to verify that the network models multiple risk functions in a population based on the specific treatment a patient undergoes. Fourth, we demonstrate how DeepSurv provides treatment recommendations and show that DeepSurv's recommendations improve a population's survival rate. For more technical details on the experiments, see Appendix \ref{sec:experimental-details}.

In addition to training DeepSurv on each dataset, we run a linear CPH regression for a baseline comparison. We also fit a RSF to compare DeepSurv against a state-of-the-art nonlinear survival model. Even though we can compare the RSF's predictive accuracy to DeepSurv's, we do not measure the RSF's performance on modeling a simulated dataset's true risk function $h(x)$. This is due to the fact that the the RSF calculates the cumulative hazard function $\Lambda(t) = \int_{0}^{t}{\lambda(\tau)d\tau}$ rather than the hazard function $\lambda(t)$. 


\subsection{Evaluation}

\subsubsection*{Survival data} To evaluate the models' predictive accuracy on the survival data, we measure the concordance-index (C-index) $c$ as outlined by \cite{harrell1984regression}. The C-index is the most common metric used in survival analysis and reflects a measure of how well a model predicts the ordering of patients' death times. For context, a $c = 0.5$ is the average C-index of a random model, whereas $c = 1$ is a perfect ranking of death times. We perform bootstrapping \citep{efron1994introduction} and sample the test set with replacement to obtain confidence intervals. 


\def \linLinearCI {\num[round-precision=6]{0.77367652681577692}} 
\def \linLinearConfd {(\num{0.77209236468783626},\num{0.77526068894371758})}
\def \linLinearMSE {\num{20.196880513197367}}
\def \deepLinearCI {\bfseries \num[round-precision=6]{0.77401880795712086}}
\def \deepLinearConfd {(\num{0.77241114821791823},\num{0.77562646769632348})}
\def \deepLinearMSE {\num{0.12577872}}
\def \rsfLinearCI {\num[round-precision=6]{0.76492506494077972}}
\def \rsfLinearConfd {(\num{0.76348517896413348},\num{0.76636495091742596})}

\def \linGaussCI {\num[round-precision=6]{0.50695098128314975}}
\def \linGaussConfd { (\num{0.50465928323923837},\num{0.50924267932706113})}

\def \deepGaussCI {\bfseries \num[round-precision=6]{0.64890227041967163}}
\def \deepGaussConfd { (\num{0.64693985541102927},  \num{0.6505651355760933})}

\def \rsfGaussCI {\num[round-precision=6]{0.64553975171239619}}
\def \rsfGaussConfd { (\num{0.64346449347896961},\num{0.64761500994582277})}

\def \linGaussMSE {\num{1.3694924484097308}}
\def \deepGaussMSE {\num{0.13908528}}

\def \whasCensor {42.12 }
\def \whasDeathMed {516.0 days}

\def \linWHASCI {\num[round-precision=6]{0.81762046541155842}}
\def \linWHASConfd {(\num{0.8143033848667216}, \num{0.82093754595639523})}

\def \deepWHASCI {\num[round-precision=6]{0.86261955265087253}}
\def \deepWHASConfd {(\num{0.85919963892050433},\num{0.86603946638124074}) }

\def \rsfWHASCI {\bfseries \num[round-precision=6]{0.89362275539520808}}
\def \rsfWHASConfd {(\num{0.89092438964609022},\num{0.89632112114432594}) }

\def \subsetWHASCI {\num{0.84308546084361791}}
\def \subsetWHASConfd {(95\% CI: \num{0.8399011692650703},\num{0.84626975242216551})}


\def \linSupportCI {\num[round-precision=6]{0.58286965205786712}}
\def \linSupportConfd { (\num{0.58106310155283947},\num{0.58467620256289476})}

\def \deepSupportCI {\bfseries \num[round-precision=6]{0.61830789482825887}}
\def \deepSupportConfd { (\num{0.61648595892083868},\num{0.62012983073567907})}

\def \rsfSupportCI {\num[round-precision=6]{0.61302156155903031}}
\def \rsfSupportConfd { (\num{0.61142695535907976},\num{0.61461616775898087})}

\def \SupportDeathMed {58 days}
\def \SupportCensor {68.10 }
\def \numNetworkFactors {14 }
\def \numPatients {9,105 }

\def \linMetabricCI {\num[round-precision=6]{0.630617604184596}}
\def \linMetabricConfd {(\num{0.62656489364171453},\num{0.63467031472747748})}

\def \deepMetabricCI {\bfseries \num[round-precision=6]{0.64337362220603711}}
\def \deepMetabricConfd {(\num{0.63949927948461494},\num{0.64724796492745929})}

\def \rsfMetabricCI {\num[round-precision=6]{0.62433108496169643}}
\def \rsfMetabricConfd {(\num{0.62003966467591687},\num{0.62862250524747598})}

\def \metabricDeathMed {116 months}
\def \metabricCensor {57.72 percent }
\def \numMetabricPatients {1,980 }

\def \linSimRecCI {\num[round-precision=6]{0.48154017856805986}}
\def \linearSimRecConfd {(\num{0.47962550696225276},\num{0.48345485017386697})}

\def \deepSimRecCI {\bfseries \num[round-precision=6]{0.58277380302993564}}
\def \deepSimRecConfd {(\num{0.58040967092789619},\num{0.5851379351319751})}

\def \rsfSimRecCI {\num[round-precision=6]{0.56987023981100404}}
\def \rsfSimRecConfd {(\num{0.56767746544888209},\num{0.57206301417312599})}

\def \deepMedianDeathSimRec {\bfseries \num{4.0688419}}
\def \deepMedianDeathAntiSimRec {\bfseries \num{2.8271899}}

\def \rsfMedianDeathSimRec {\num{3.1162324}}
\def \rsfMedianDeathAntiSimRec {\num{3.6249108}}
 
\def \deepSurvivalSimCurvePvalue {1$\mathrm{e}{-5}$}
\def \rsfSurvivalSimCurvePvalue {\num[round-precision=6]{0.05536}}

\def \linRottCI {\num[round-precision=6]{0.65775014921700059}}
\def \linRottConf {(\num{0.65409056084924277}, \num{0.66140973758475841})}

\def \deepRottCI {\bfseries \num[round-precision=6]{0.66840201171649061}}
\def \deepRottConf {(\num{0.66545919099744943},\num{0.67134483243553178})}

\def \rsfRottCI {\num[round-precision=6]{0.65118980785489233}}
\def \rsfRottConf {(\num{0.64793025589639708}, \num{0.65444935981338759})}

\def \deepMedianDeathRec {\bfseries \num{40.098564}}
\def \deepMedianDeathActual {\bfseries \num{31.770021}}

\def \rsfMedianDeathRec {\num{36.566734}}
\def \rsfMedianDeathActual {\num{32.394249}}
 
\def \deepSurvivalCurvePvalue {\num[round-precision=6]{0.00343}}
\def \rsfSurvivalCurvePvalue {\num[round-precision=6]{0.08334}}

{\renewcommand{\arraystretch}{1.2}
\begin{table}[h]
\caption{Experimental Results for All Experiments: C-index (95\% Confidence Interval)}
\label{table:results}
\begin{tabular}{ |p{2.5cm}||p{3.2cm}|p{3.2cm}|p{3.2cm}| }
 \hline
 Experiment & CPH & DeepSurv & RSF\\
 \hline
 Simulated Linear  & \linLinearCI \newline \linLinearConfd & \deepLinearCI \newline \deepLinearConfd & \rsfLinearCI \newline \rsfLinearConfd \\
 \hline
 Simulated Nonlinear & \linGaussCI \newline \linGaussConfd  &  \deepGaussCI \newline \deepGaussConfd   &  \rsfGaussCI \newline \rsfGaussConfd\\
 \hline
 WHAS & \linWHASCI \newline \linWHASConfd & \deepWHASCI \newline \deepWHASConfd & \rsfWHASCI \newline \rsfWHASConfd \\
 \hline
 SUPPORT    & \linSupportCI \newline \linSupportConfd & \deepSupportCI \newline \deepSupportConfd &  \rsfSupportCI \newline \rsfSupportConfd \\
 \hline
  METABRIC & \linMetabricCI \newline \linMetabricConfd  &  \deepMetabricCI \newline \deepMetabricConfd & \rsfMetabricCI \newline \rsfMetabricConfd \\
 \hline
 Simulated Treatment & \linSimRecCI \newline \linearSimRecConfd &  \deepSimRecCI \newline \deepSimRecConfd &  \rsfSimRecCI \newline \rsfSimRecConfd \\
 \hline
 Rotterdam \& GBSG & \linRottCI \newline \linRottConf  &  \deepRottCI \newline \deepRottConf & \rsfRottCI \newline \rsfRottConf \\
 \hline
\end{tabular}
\end{table}

\subsubsection*{Treatment recommendations} 
We determine the recommended treatment for each patient in the test set using DeepSurv and the RSF. We do not calculate the recommended treatment for CPH; without preselected treatment-interaction terms, the CPH model will compute a constant recommender function and recommend the same treatment option for all patients. This would effectively be comparing the survival rates between the control and experimental groups. DeepSurv and the RSF are capable of predicting an individual's risk per treatment because each computes relevant interaction terms. For DeepSurv, we choose the recommended treatment by calculating the recommender function (Equation \ref{eq:rec(x)}). Because the RSF predicts a cumulative hazard for each patient, we choose the treatment with the minimum cumulative hazard. 

Once we determine the recommended treatment, we identify two subsets of patients: those whose treatment group aligns with the model's recommended treatment (Recommendation) and those who do not undergo the recommended treatment (Anti-Recommendation). We calculate the median survival time of each subset to determine if a model's treatment recommendations increase the survival rate of the patients. We then perform a log-rank test to validate whether the difference between the two subsets is significant.

\begin{table}[h]
\caption{Experimental Results for Treatment Recommendations: Median Survival Time (months)}
\label{table:treatment}
\begin{tabular}{|p{2.5cm}||p{2.25cm}|p{2.25cm}|p{2.25cm}|p{2.25cm} | }
\hline
\multirow{2}{*}{Experiment} & \multicolumn{2}{c|}{DeepSurv} & \multicolumn{2}{c|}{RSF} \\
\cline{2-5}
 & Rec & Anti-Rec & Rec & Anti-Rec\\
\hline
Simulated & \deepMedianDeathSimRec & \deepMedianDeathAntiSimRec & \rsfMedianDeathSimRec & \rsfMedianDeathAntiSimRec \\
\hline
Rotterdam \& GBSG & \deepMedianDeathRec & \deepMedianDeathActual & \rsfMedianDeathRec & \rsfMedianDeathActual \\
\hline
\end{tabular}
\end{table}

\subsection{Simulated survival data} \label{sec:simulated}

In this section, we perform two experiments with simulated survival data: one with a linear risk function and one with a nonlinear (Gaussian) risk function. The advantage of using simulated datasets is that we can ascertain whether DeepSurv can successfully model the true risk function instead of overfitting random noise.

\def \numSimExamples {4000, 1000, 1000}

For each experiment, we generate a training, validation, and testing set of $N = \numSimExamples$ observations respectively. Each observation represents a patient vector with $d = 10$ covariates, each drawn from a uniform distribution on $[-1,1)$. We generate the death time $T$ according to an exponential Cox model \citep{austin2012generating}:
\begin{equation}
T \sim \mathrm{Exp} ( \lambda(t; x) ) = \mathrm{Exp} ( \lambda_0 \cdot e^{h(x)} ) 
\end{equation}
Details of the simulated data generation are found in Appendix \ref{appendix:simulated}.

\def \avgObserved {90 }

In both experiments, the risk function $h(x)$ only depends on two of the ten covariates, and we demonstrate that DeepSurv discerns the relevant covariates from the noise. We then choose a censoring time to represent the `end of study' such that about \avgObserved percent of the patients have an observed event in the dataset. 

\subsubsection{Linear risk experiment}
We first simulate patients to have a linear risk function for $x \in \mathbb{R}^d$ so that the linear proportional hazards assumption holds true:
\begin{equation} \label{eq:risk-linear}
 h(x) = x_{0} + 2x_{1}.
 \end{equation}
Because the linear proportional hazards assumption holds true, we expect the linear CPH to accurately model the risk function in Equation \ref{eq:risk-linear}. 

Our results (see Table \ref{table:results}) demonstrate that DeepSurv performs as well as the standard linear Cox regression and better than RSF in predictive ability.

\begin{figure}[!tpb]
	\centering
	\begin{tabular}{ccc}
	\multirow{2}{*}{
	\begin{subfigure}{0.3\columnwidth} 		
		\centering
		\includegraphics[width=\columnwidth]{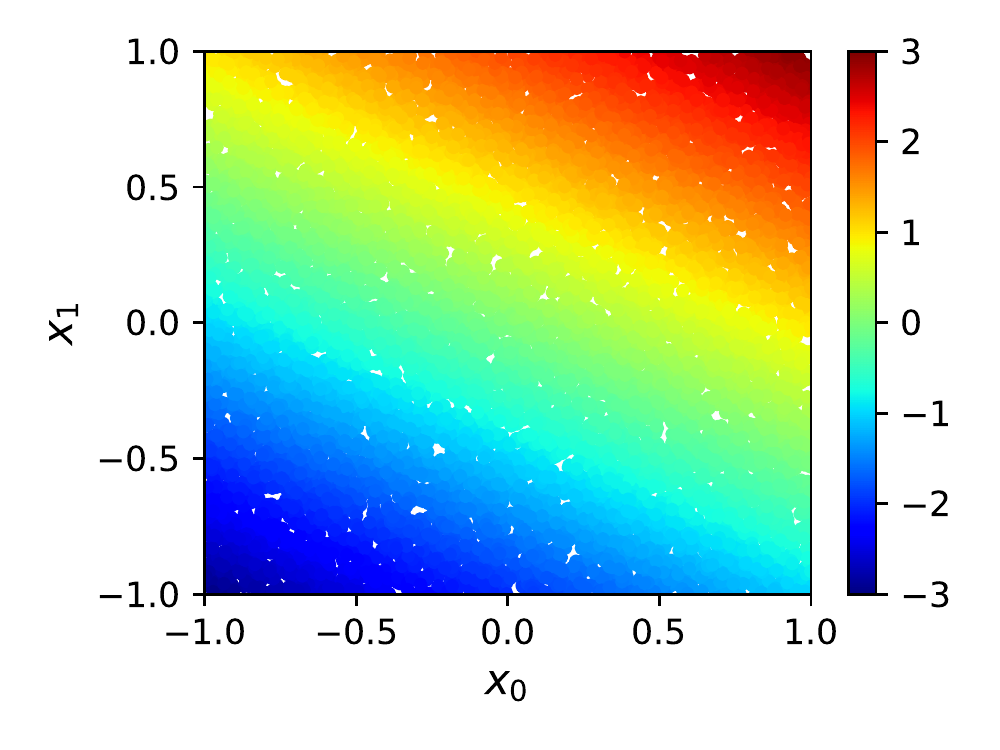}
		\caption{True $h(x)$}
		 \label{fig:linear_true}
	\end{subfigure}
	}  & 
	\begin{subfigure}{0.3\columnwidth} 		
		\centering
		\includegraphics[width=\columnwidth]{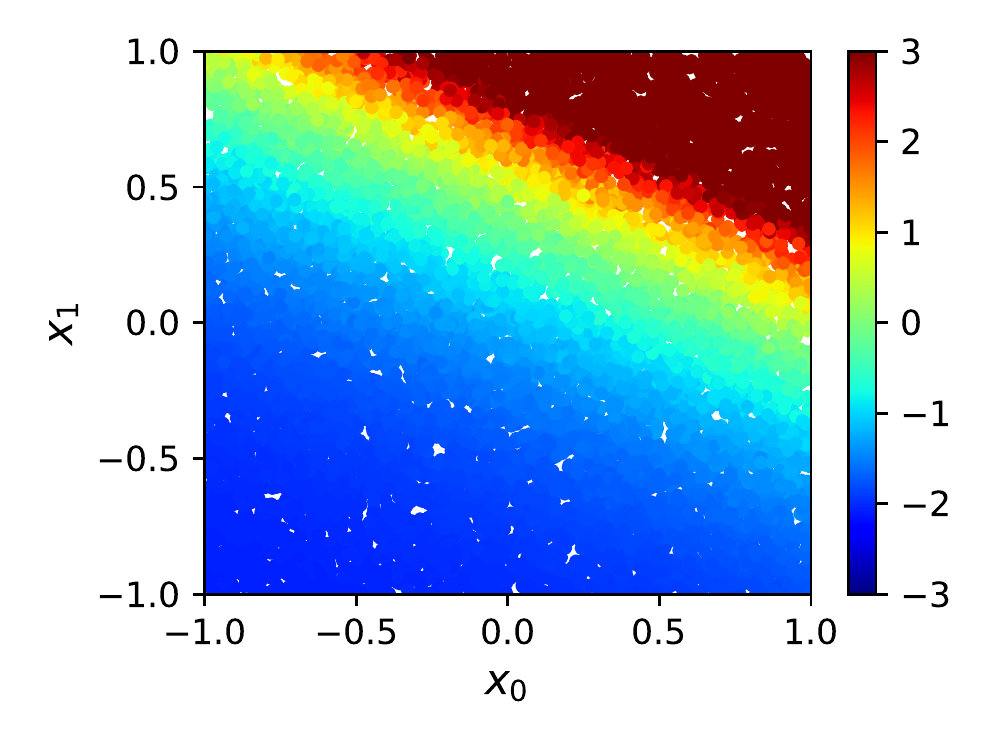}
		\caption{CPH $\hat{h}_\beta(x)$}
		 \label{fig:linear_CPH}
	\end{subfigure} & 
	\begin{subfigure}{0.3\columnwidth} 		
		\centering
		\includegraphics[width=\columnwidth]{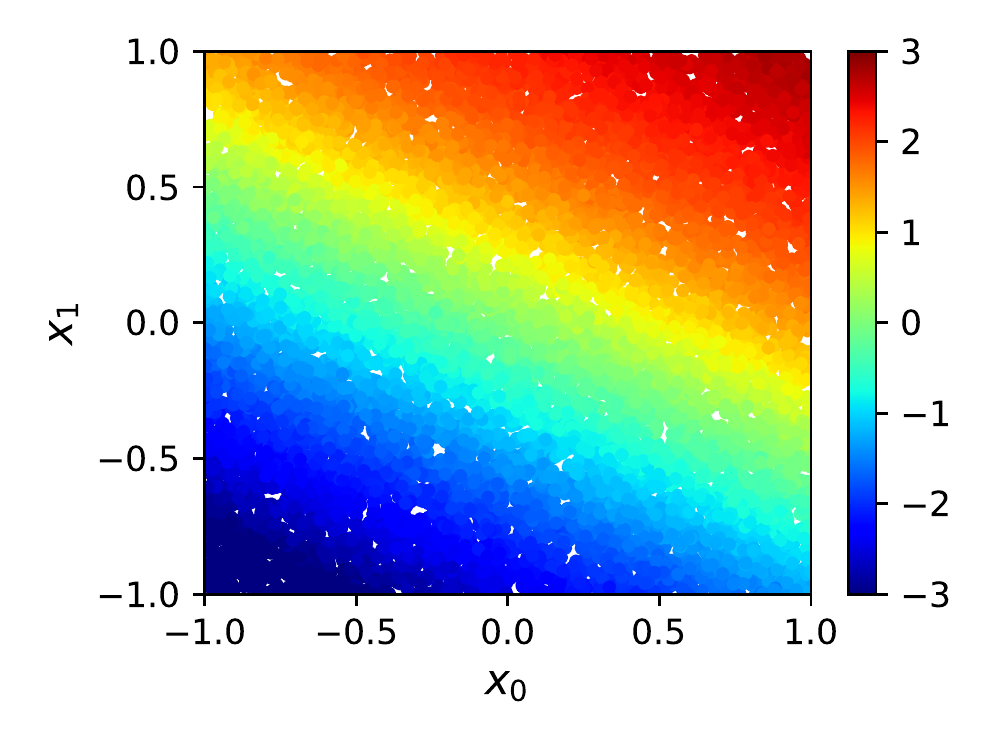}
		\caption{DeepSurv $\hat{h}_\theta(x)$}
		 \label{fig:linear_DCPH}
	\end{subfigure} \\ 
	&
	\begin{subfigure}{0.3\columnwidth} 		
		\centering
		\includegraphics[width=\columnwidth]{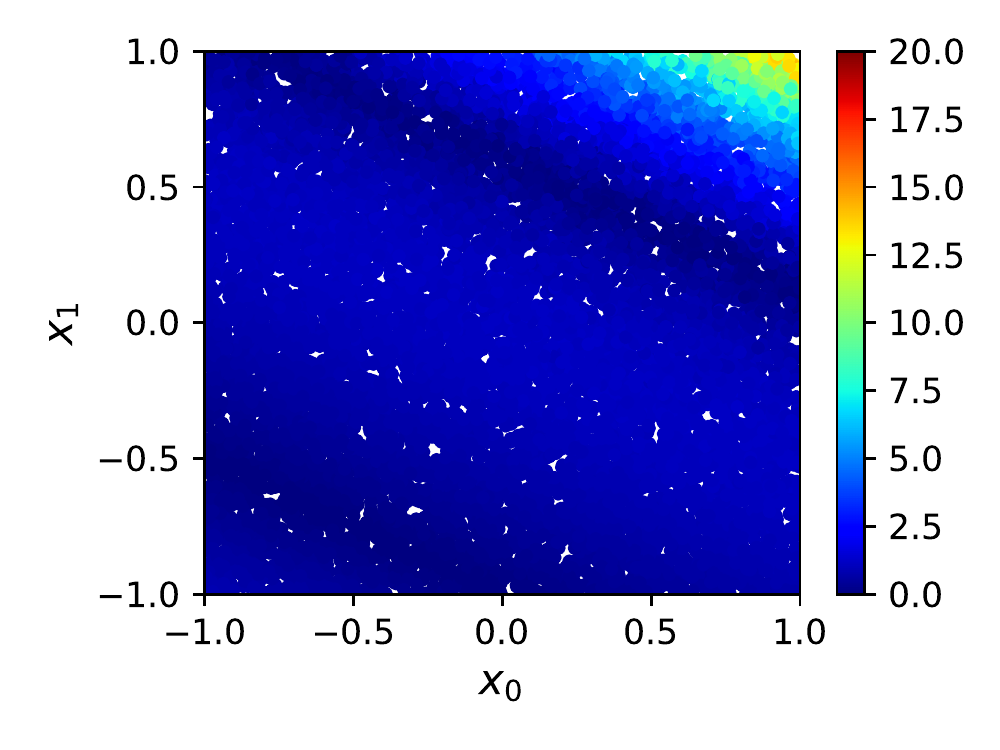}
		\caption{$|h(x) - \hat{h}_\beta(x)|$}
		 \label{fig:CPH_error}
	\end{subfigure} & 
		\begin{subfigure}{0.3\columnwidth} 		
		\centering
		\includegraphics[width=\columnwidth]{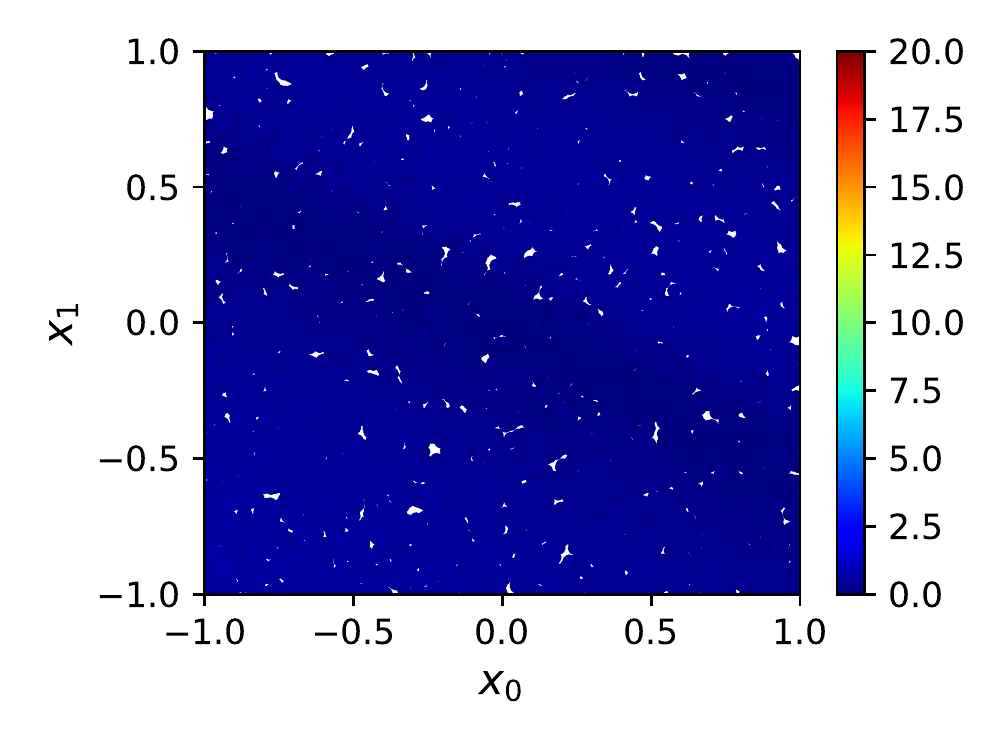}
		\caption{$|h(x) - \hat{h}_\theta(x)|$}
		 \label{fig:DeepSurv_error}
	\end{subfigure}
\end{tabular}
\caption{Predicted risk surfaces and errors for the simulated survival data with linear risk function with respect to a patient's covariates $x_0$ and $x_1$. \ref{fig:linear_true} The true risk $h(x) = x_0 + 2x_1$ for each patient. \ref{fig:linear_CPH} The predicted risk surface of $\hat{h}_\beta(x)$ from the linear CPH model parameterized by $\beta$. \ref{fig:linear_DCPH} The output of DeepSurv $\hat{h}_\theta(x)$ predicts a patient's risk. \ref{fig:CPH_error} The absolute error between true risk $h(x)$ and CPH's predicted risk $\hat{h}_\beta(x)$. \ref{fig:DeepSurv_error} The absolute error between true risk $h(x)$ and DeepSurv's predicted risk $\hat{h}_\theta(x)$.}
\label{fig:linear}
\end{figure}

Figure \ref{fig:linear} demonstrates how DeepSurv more accurately models the risk function compared to the linear CPH. Figure \ref{fig:linear_true} plots the true risk function $h(x)$ for all patients in the test set. As shown in Figure \ref{fig:linear_CPH}, the CPH's estimated risk function $\hat{h}_\beta(x)$ does not perfectly model the true risk for a patient. In contrast, as shown in Figure \ref{fig:linear_DCPH}, DeepSurv better estimates the true risk function. 

To quantify these differences, Figures \ref{fig:CPH_error} and \ref{fig:DeepSurv_error} show that the CPH's estimated risk has a significantly larger absolute error than that of DeepSurv, specifically for patients with a high positive risk. We calculate the mean-squared-error (MSE) between a model's predicted risk and the true risk values. The MSEs of CPH and DeepSurv are \linLinearMSE \: and \deepLinearMSE, respectively. Even though DeepSurv and CPH have similar predictive abilities, this demonstrates that DeepSurv is superior than the CPH at modeling the true risk function of the population. 

\subsubsection{Nonlinear risk experiment} \label{sec:nonlinear} 

We set the risk function to be a Gaussian with $\lambda_{\max} = 5.0$ and a scale factor of $r = 0.5$:
\begin{equation}
\label{eq:gaussian}
h(x) = \log (\lambda_{\max}) \: \exp \left( -{\frac{x_{0}^2 + x_{1}^2}{2 r^2}} \right) 
\end{equation} 
The surface of the risk function is depicted in \ref{fig:gauss_true}. Because this risk function is nonlinear, we do not expect the CPH to predict the risk function properly without adding quadratic terms of the covariates to the model. We expect DeepSurv to reconstruct the Gaussian risk function and successfully predict a patient's risk. Lastly, we expect the RSF and DeepSurv to accurately rank the order of patient's deaths.

The CI results in Table \ref{table:results} shows that DeepSurv outperforms the linear CPH and predicts as well as the RSF. In addition, DeepSurv correctly learns nonlinear relationships between a patient's covariates and their risk. As shown in Figure \ref{fig:gauss}, DeepSurv is more successful than the linear CPH in modeling the true risk function. Figure \ref{fig:gauss_lin} demonstrates that the linear CPH regression fails to determine the first two covariates as significant. The CPH has a C-index of \linGaussCI, which is equivalent to the performance of randomly ranking death times. Meanwhile, Figure \ref{fig:gauss_pred} demonstrates that DeepSurv reconstructs the Gaussian relationship between the first two covariates and a patient's risk. 

\begin{figure}[!tpb]
	\centering
	\begin{subfigure}[b]{0.3\columnwidth}
		\centering
		\includegraphics[width=\columnwidth]{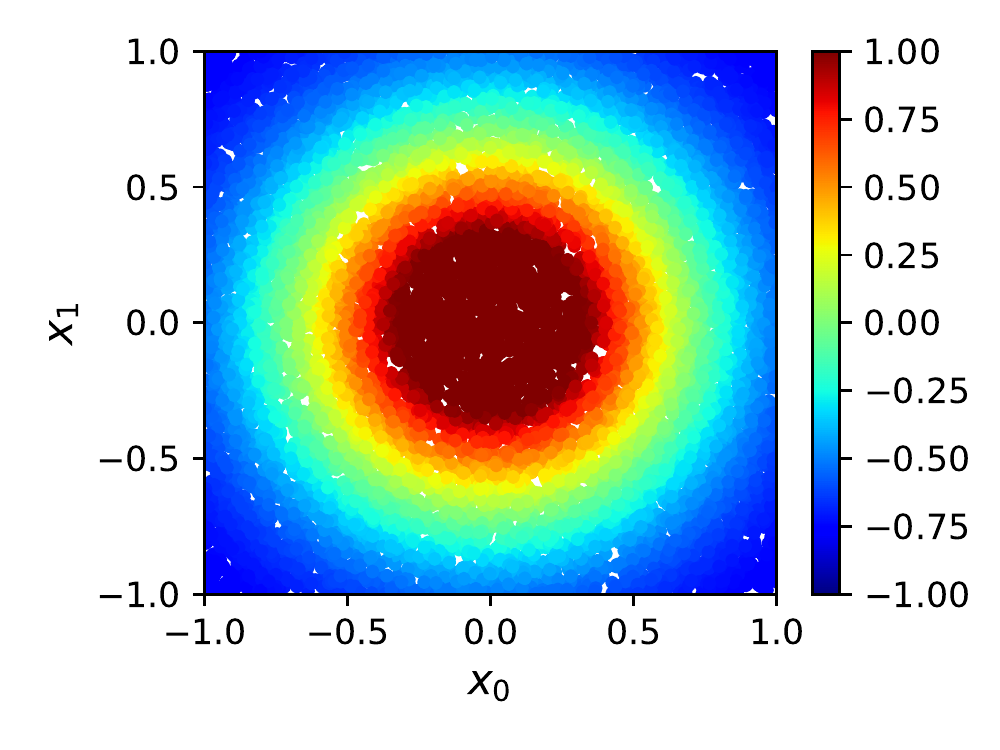}
		\caption{True $h(x)$}
		\label{fig:gauss_true}
	\end{subfigure}
	\:
	\begin{subfigure}[b]{0.3\columnwidth}
		\centering
		\includegraphics[width=\columnwidth]{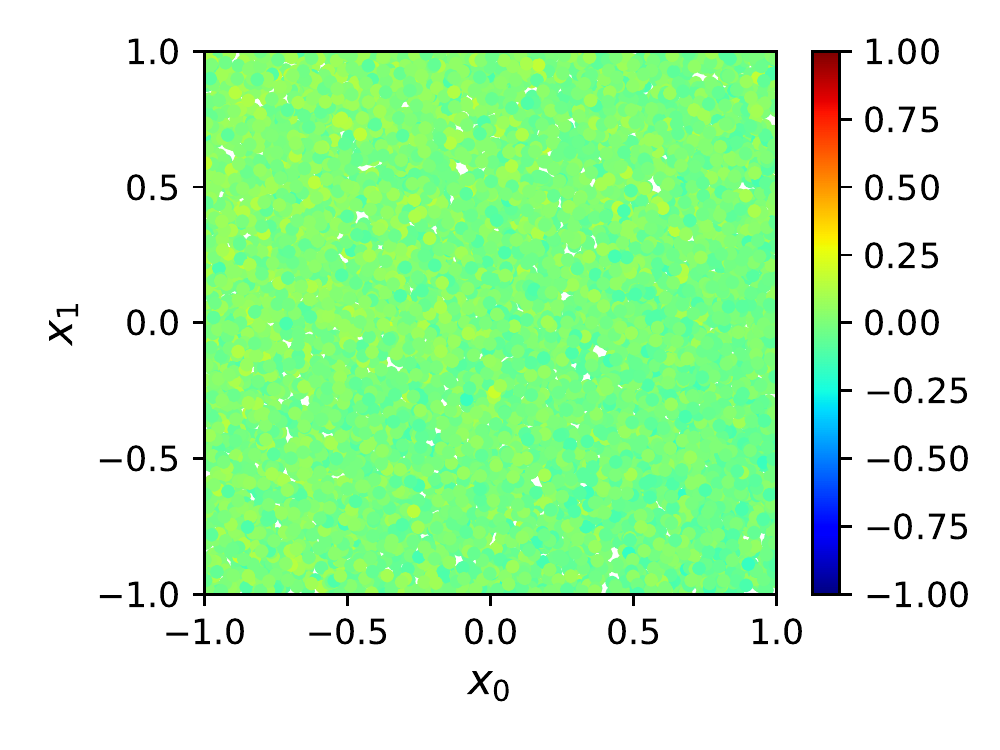}
		\caption{CPH $\hat{h}_\beta(x)$}
		\label{fig:gauss_lin}
	\end{subfigure}
		\:
	\begin{subfigure}[b]{0.3\columnwidth}
		\centering
		\includegraphics[width=\columnwidth]{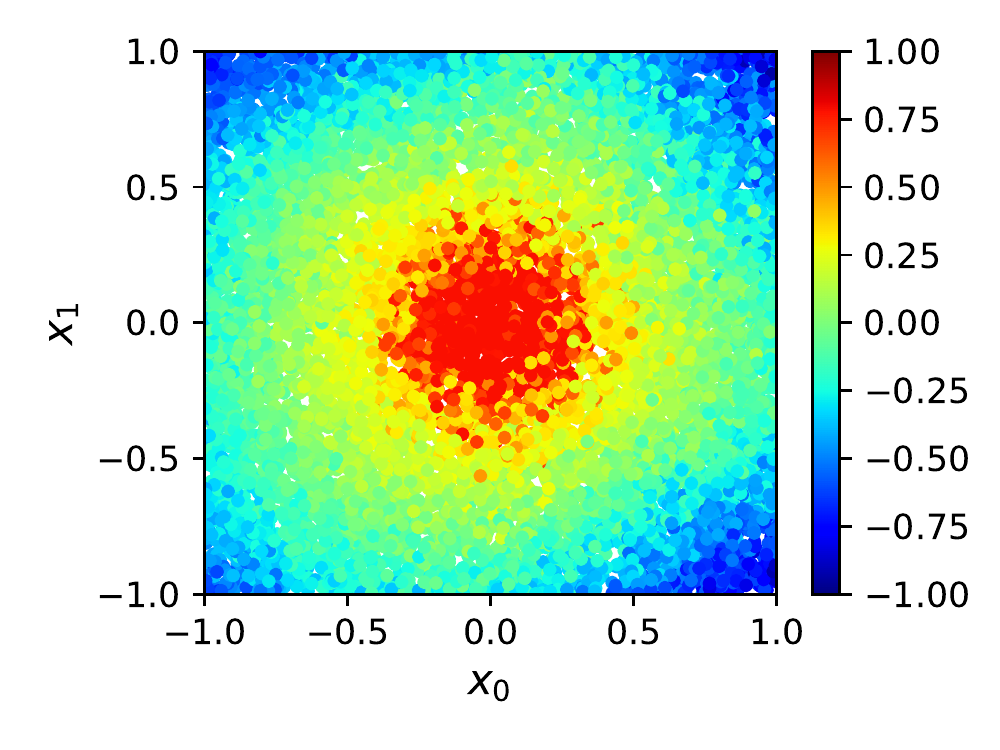}
		\caption{DeepSurv $\hat{h}_\theta(x)$}
		\label{fig:gauss_pred}
	\end{subfigure}
	\caption {Risk surfaces of the nonlinear test set with respect to patient's covariates $x_0$ and $x_1$. \ref{fig:gauss_true} The calculated true risk $h(x)$ (Equation \ref{eq:gaussian}) for each patient. \ref{fig:gauss_lin} The predicted risk surface of $\hat{h}_\beta(x)$ from the linear CPH model parameterized on $\beta$. The linear CPH predicts a constant risk. \ref{fig:gauss_pred} The output of DeepSurv $\hat{h}_\theta(x)$ is the estimated risk function.	\vspace{-2ex}}
	\label{fig:gauss}
\end{figure}

\subsection{Real survival data experiments}
We compare the performance of the CPH and DeepSurv on three datasets from real studies: the Worcester Heart Attack Study (WHAS), the Study to Understand Prognoses Preferences Outcomes and Risks of Treatment (SUPPORT), and The Molecular Taxonomy of Breast Cancer International Consortium (METABRIC). Because previous research shows that neural networks do not outperform the CPH, our goal is to demonstrate that DeepSurv does indeed have state-of-the-art predictive ability in practice on real survival datasets.

\subsubsection{Worcester Heart Attack Study (WHAS)}

The Worcester Heart Attack Study (WHAS) investigates the effects of a patient's factors on acute myocardial infraction (MI) survival \citep{hosmer9780471754992}. The dataset consists of 1,638 observations and 5 features: age, sex, body-mass-index (BMI), left heart failure complications (CHF), and order of MI (MIORD). We reserve 20 percent of the dataset as a testing set. A total of \whasCensor percent of patients died during the survey with a median death time of \whasDeathMed. As shown in Table \ref{table:results}, DeepSurv outperforms the CPH; however, the RSF outperforms DeepSurv.  


\subsection{Study to Understand Prognoses Preferences Outcomes and Risks of Treatment (SUPPORT)}

The Study to Understand Prognoses Preferences Outcomes and Risks of Treatment (SUPPORT) is a larger study that researches the survival time of seriously ill hospitalized adults \citep{knaus1995support}. The dataset consists of \numPatients patients and \numNetworkFactors features for which almost all patients have observed entries (age, sex, race, number of comorbidities, presence of diabetes, presence of dementia, presence of cancer, mean arterial blood pressure, heart rate, respiration rate, temperature, white blood cell count, serum's sodium, and serum's creatinine). We drop patients with any missing features and reserve 20 percent of the dataset as a testing set. A total of \SupportCensor percent of patients died during the survey with a median death time of \SupportDeathMed. 

As shown in Table \ref{table:results}, DeepSurv performs as well as the RSF and better than the CPH with a larger study. This validates DeepSurv's ability to predict the ranking of patient's risks on real survival data.

\subsubsection{Molecular Taxonomy of Breast Cancer International Consortium (METABRIC)}
The Molecular Taxonomy of Breast Cancer International Consortium (METABRIC) uses gene and protein expression profiles to determine new breast cancer subgroups in order to help physicians provide better treatment recommendations. 

The METABRIC dataset consists of gene expression data and clinical features for \numMetabricPatients patients, and \metabricCensor have an observed death due to breast cancer with a median survival time of \metabricDeathMed\: \citep{curtis2012genomic}. We prepare the dataset in line with the Immunohistochemical 4 plus Clinical (IHC4+C) test, which is a common prognostic tool for evaluating treatment options for breast cancer patients \citep{lakhanpal2016ihc4}. We join the 4 gene indicators (\textit{MKI67, EGFR, PGR, and ERBB2}) with the a patient's clinical features (hormone treatment indicator, radiotherapy indicator, chemotherapy indicator, ER-positive indicator, age at diagnosis). We then reserved 20 percent of the patients as the test set.

Table \ref{table:results} shows that DeepSurv performs better than both the CPH and RSF. This result demonstrates not only DeepSurv's ability to model the risk effects of gene expression data but also shows the potential for future research of DeepSurv as a comparable prognostic tool to common medical tests such as the IHC4+C. 


\subsection{Treatment recommender system experiments}

In this section, we perform two experiments to demonstrate the effectiveness of DeepSurv's treatment recommender system. First, we simulate treatment data by including an additional covariate to the simulated data from Section \ref{sec:nonlinear}. Second, after demonstrating DeepSurv's modeling and recommendation capabilities, we apply the recommender system to a real dataset used to study the effects of hormone treatment on breast cancer patients. We show that DeepSurv can successfully provide personalized treatment recommendations. We conclude that if all patients follow the network's recommended treatment options, we would gain a significant increase in patients' lifespans.

\subsubsection{Simulated treatment data}

We uniformly assign a treatment group $\tau \in \{ 0, 1 \}$ to each simulated patient in the dataset. All of the patients in group $\tau =0$ were `unaffected' by the treatment (e.g. given a placebo) and have a constant risk function $h_0(x)$. The other group $\tau =1$ is prescribed a treatment with Gaussian effects (Equation \ref{eq:gaussian}) and has a risk function $h_1(x)$ with $\lambda_{\max}=10$ and $r=0.5$. 

Figure \ref{fig:SimRecommender} illustrates the network's success in modeling both treatments' risk functions for patients. Figure \ref{fig:SimRec_true} plots the true risk distribution $h(x)$. As expected, Figure \ref{fig:SimRec_group0} shows that the network models a constant risk for a patient in treatment $\tau=0$, independent of a patient's covariates. Figure \ref{fig:SimRec_group1} shows how DeepSurv models the Gaussian effects of a patient's covariates on their treatment risk. To further quantify these results, Table \ref{table:results} shows that DeepSurv has the largest concordance index. Because the network accurately reconstructs the risk function, we expect that it will provide accurate treatment recommendations for new patients.

\begin{figure}[h]
        \centering
        \begin{subfigure}[b]{0.3\columnwidth}
            \centering
            \includegraphics[width=\columnwidth]{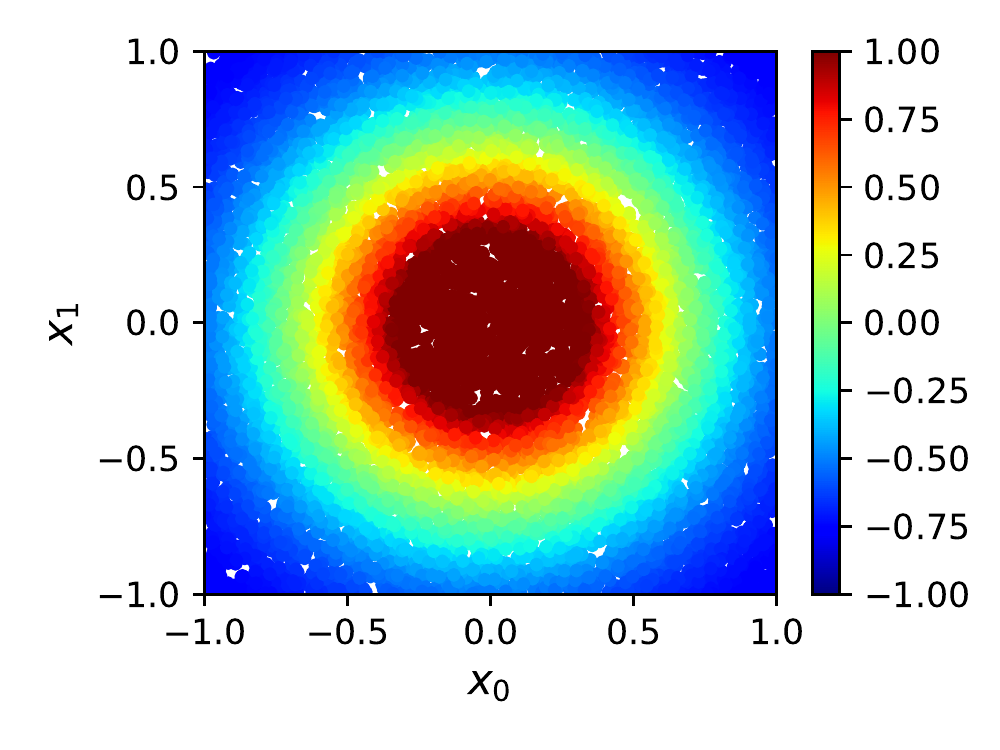}
            \caption[]
            {{True $h(x)$}}    
            \label{fig:SimRec_true}
        \end{subfigure}
        \:
        \begin{subfigure}[b]{0.3\columnwidth}
            \centering 
            \includegraphics[width=\columnwidth]{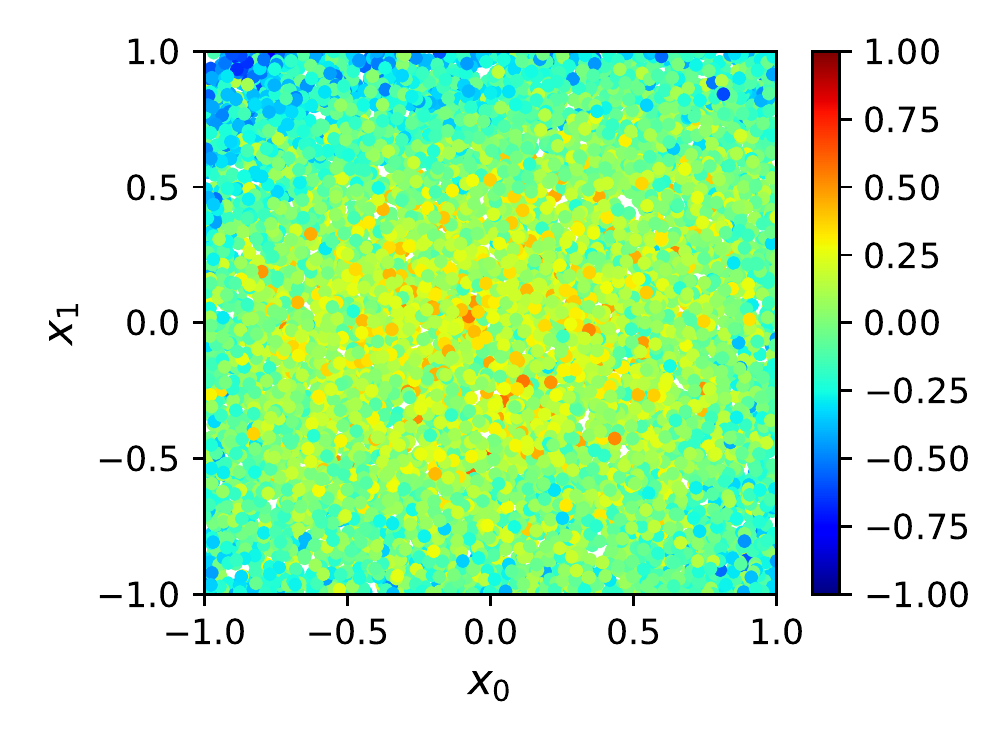}
            \caption[]%
            {{DeepSurv $\hat{h}_0(x)$}}    
            \label{fig:SimRec_group0}
        \end{subfigure}
        \:
        \begin{subfigure}[b]{0.3\columnwidth}
            \centering 
            \includegraphics[width=\columnwidth]{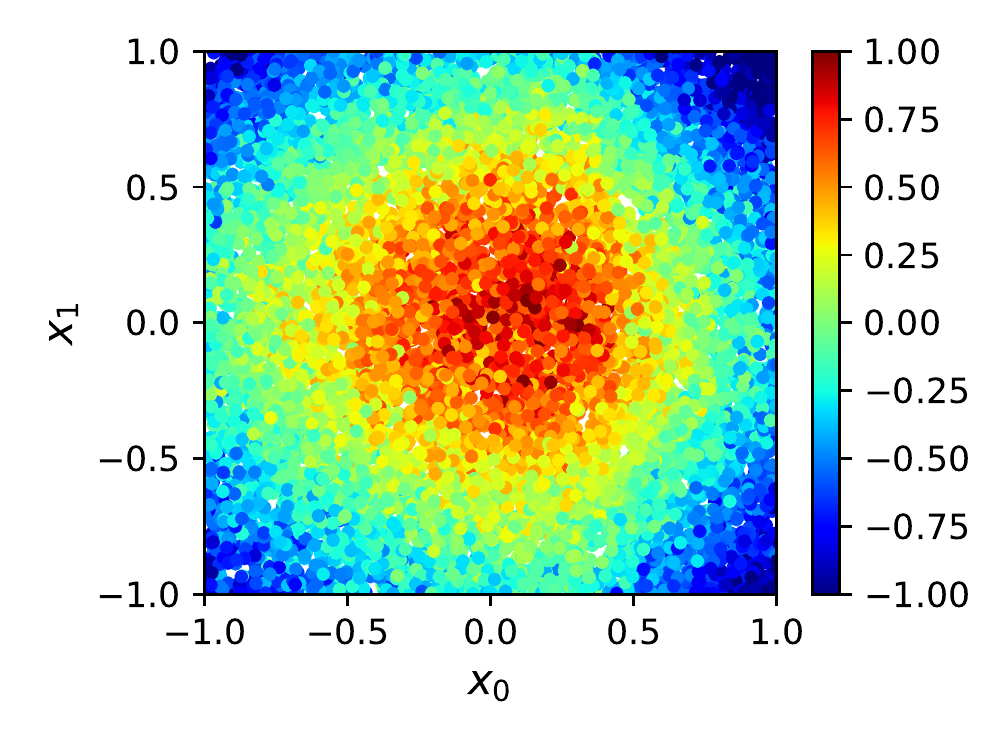}
            \caption[]%
            {{DeepSurv $\hat{h}_1(x)$}}    
            \label{fig:SimRec_group1}
        \end{subfigure}
        \caption
        {Treatment Risk Surfaces as a function of a patient's relevant covariates $x_0$ and $x_1$. \ref{fig:SimRec_true} The true risk $h_1(x)$ if all patients in the test set were given treatment $\tau=1$. We then manually set all treatment groups to either $\tau = 0$ or $\tau = 1$. \ref{fig:SimRec_group0} The predicted risk $\hat{h}_0(x)$ for patients with treatment group $\tau = 0$. \ref{fig:SimRec_group1} The network's predicted risk $\hat{h}_1(x)$ for patients in treatment group $\tau = 1$.} 
        \label{fig:SimRecommender}
\end{figure}

\begin{figure}[h]
	\vspace{-2ex}
	\centering
	\begin{subfigure}[b]{\columnwidth}
		\centering
		\includegraphics[width=\columnwidth]{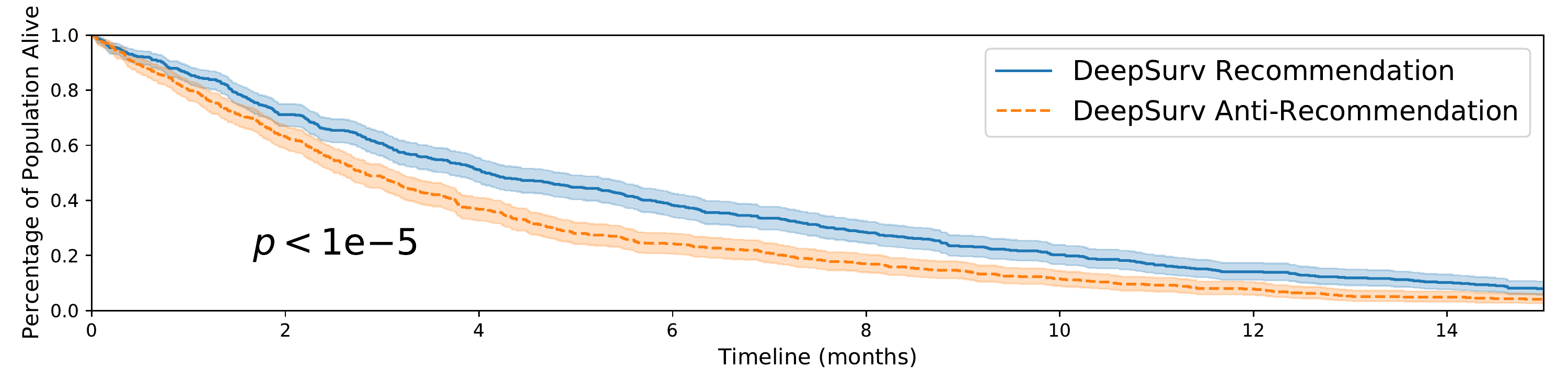}
		\caption{Effect of DeepSurv's Treatment Recommendations (Simulated Data)}
		\label{fig:deep_sim_rec}
	\end{subfigure}
	\begin{subfigure}[b]{\columnwidth}
		\centering
		\includegraphics[width=\columnwidth]{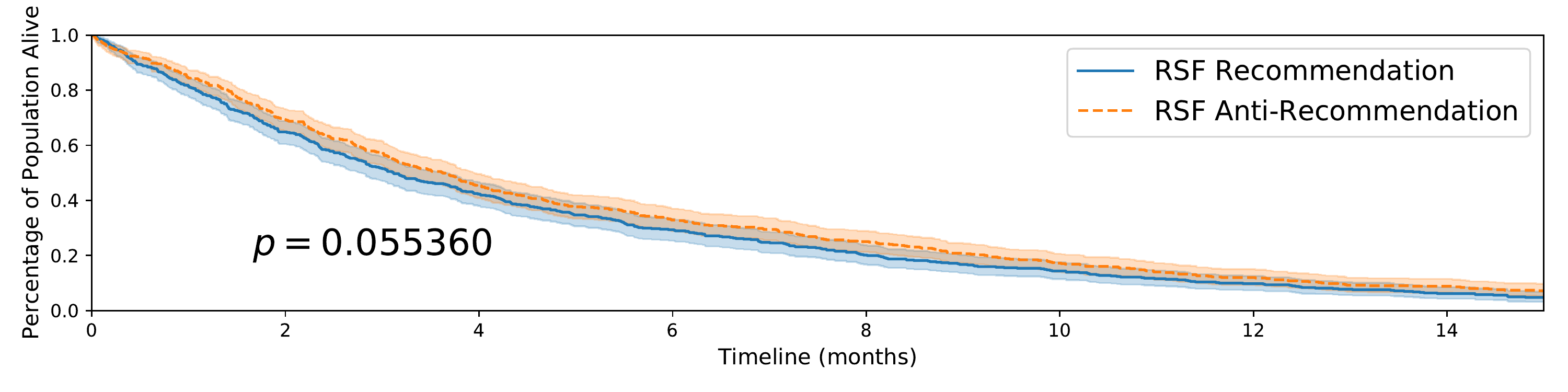}
		\caption{Effect of RSF's Treatment Recommendations (Simulated Data)}
		\label{fig:rsf_sim_rec}
	\end{subfigure}
	\caption {Kaplan-Meier estimated survival curves with confidence intervals ($\alpha = .05$) for the patients who were given the treatment concordant with a method's recommended treatment (Recommendation) and the subset of patients who were not (Anti-Recommendation). We perform a log-rank test to validate the significance between each set of survival curves.	\vspace{-2ex}}
	\label{fig:sim-recommender-experiment}
\end{figure}

In Figure \ref{fig:sim-recommender-experiment}, we plot the Kaplan-Meier survival curves for both the Recommendation and Anti-Recommendation subset for each method. Figure \ref{fig:deep_sim_rec} shows that the survival curve for the Recommendation subset is shifted to the right, which signifies an increase in survival time for the population following DeepSurv's recommendations. This is further quantified by the median survival times summarized in Table \ref{table:treatment}. The p-value of DeepSurv's recommendations is less than \deepSurvivalSimCurvePvalue, and we can reject the null hypothesis that DeepSurv's recommendations would not affect the population's survival time. As shown in Table \ref{table:treatment}, the subset of patients that follow RSF's recommendations have a shorter survival time than those who do not follow RSF's recommended treatment. Therefore, we could take the RSF's recommendations and provide the patients with the opposite treatment option to increase median survival time; however, Figure \ref{fig:rsf_sim_rec} shows that that improvement would not be statistically valid. While both methods of DeepSurv and RSF are able to compute treatment interaction terms, DeepSurv is more successful in recommending personalized treatments.

\subsubsection{Rotterdam \& German Breast Cancer Study Group (GBSG)}

We first train DeepSurv on breast cancer data from the Rotterdam tumor bank \citep{foekens2000urokinase}.  and construct a recommender system to provide treatment recommendations to patients from a study by the German Breast Cancer Study Group (GBSG) \citep{schumacher1994randomized}. The Rotterdam tumor bank dataset contains records for 1,546 patients with node-positive breast cancer, and nearly 90 percent of the patients have an observed death time. The testing data from the GBSG contains complete data for 686 patients (56 percent are censored) in a randomized clinical trial that studied the effects of chemotherapy and hormone treatment on survival rate. We preprocess the data as outlined by \cite{altman2000we}. 

We first validate DeepSurv's performance against the RSF and CPH baseline. We then plot the two survival curves: the survival times of those who followed the recommended treatment and those who did not. If the recommender system is effective, we expect the population with the recommended treatments to survive longer than those who did not take the recommended treatment.

\begin{figure}[h!]
	\vspace{-2ex}
	\centering
	\begin{subfigure}[b]{\columnwidth}
		\centering
		\includegraphics[width=\columnwidth]{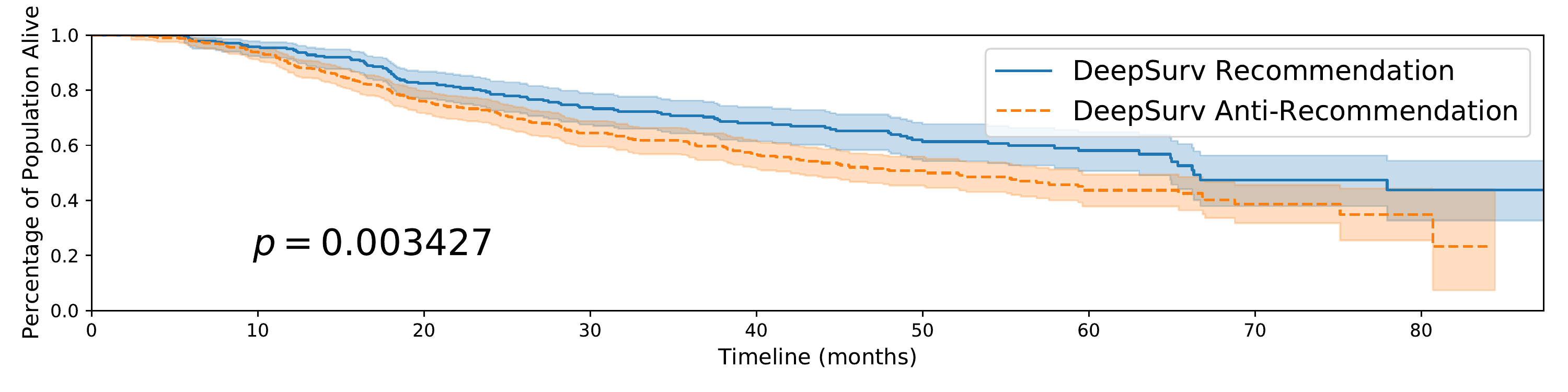}
		\caption{Effect of DeepSurv's Treatment Recommendations (GBSG)}
		\label{fig:dcph_tamox}
	\end{subfigure}
	\begin{subfigure}[b]{\columnwidth}
		\centering
		\includegraphics[width=\columnwidth]{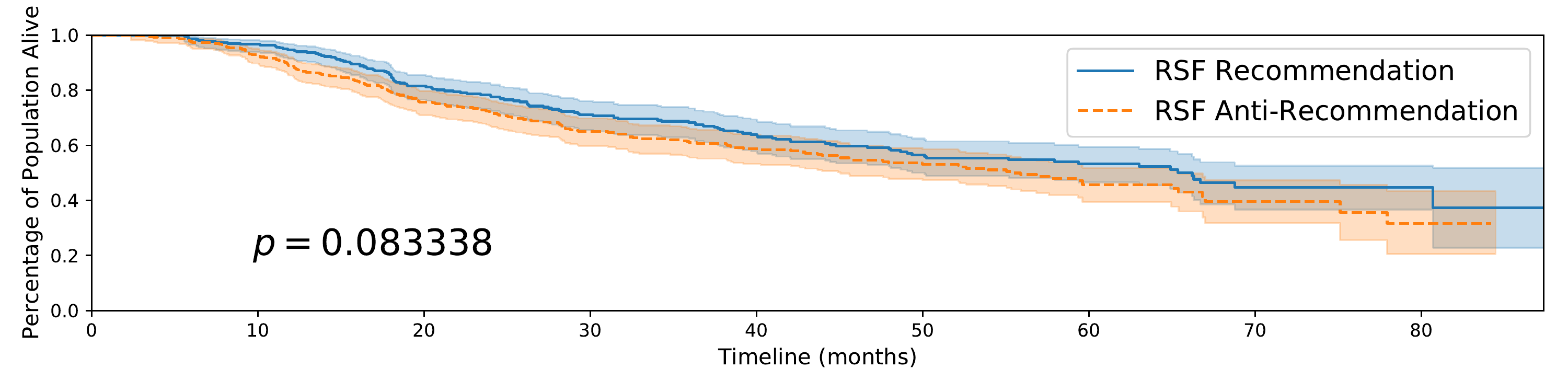}
		\caption{Effect of RSF's Treatment Recommendations (GBSG)}
		\label{fig:rsf_tamox}
	\end{subfigure}
	\caption {Kaplan-Meier estimated survival curves with confidence intervals ($\alpha = .05$) for the patients who were given the treatment concordant with a method's recommended treatment (Recommendation) and the subset of patients who were not (Anti-Recommendation). We perform a log-rank test to validate the significance between each set of survival curves.	\vspace{-2ex}}
	\label{fig:recommender-experiment}
\end{figure}

Table \ref{table:results} shows that DeepSurv provides an improved predictive ability relative to the CPH and RSF. In Figure \ref{fig:recommender-experiment}, we plot the Kaplan-Meier survival curves for both the Recommendation subset and the Anti-Recommendation subset for each method. Figure \ref{fig:dcph_tamox} shows that the survival curve for DeepSurv's Recommendation subset is statistically significant from the Anti-recommendation subset, and Table \ref{table:treatment} shows that DeepSurv's recommendations increase the median survival time of the population. Figure \ref{fig:rsf_tamox} demonstrates that RSF is unable to provide significant treatment recommendations, despite an increase in median survival times (see Table \ref{table:treatment}). The results of this experiment demonstrate not only DeepSurv's superior modeling capabilities but also validate DeepSurv's ability in providing personalized treatment recommendations on real clinical data. Moreover, we can train DeepSurv on survival data from one clinical study and transfer the learnings to provide personalized treatment recommendations to a different population of breast cancer patients.

\section{Conclusion}
\label{sec:conclusion}

In conclusion, we demonstrated that the use of deep learning in survival analysis allows for: (i) higher performance due to the flexibility of the model, and (ii) effective treatment recommendations based on the predicted effect of treatment options on an individual's risk. We validated that DeepSurv predicts patients' risk mostly as well as or better than other linear and nonlinear survival methods. We experimented on increasingly complex survival datasets and demonstrated that DeepSurv computes complex and nonlinear features without \textit{a priori} selection or domain expertise. We then demonstrated that DeepSurv is superior in predicting personalized treatment recommendations compared to the state-of-the-art survival method of random survival forests. We also released a Python module that implements DeepSurv and scripts for running reproducible experiments in Docker, see \href{https://github.com/jaredleekatzman/DeepSurv}{https://github.com/jaredleekatzman/DeepSurv} for more details. The success of DeepSurv's predictive, modeling, and recommending abilities paves the way for future research in deep neural networks and survival analysis. DeepSurv can lead to various extensions, such as the use of convolution neural networks to predict risk with medical imaging. With more research at scale, DeepSurv has the potential to supplement traditional survival analysis methods and become a standard method for medical practitioners to study and recommend personalized treatment options.

\section*{Funding}

This research was partially funded by a National Institutes of Health grant [1R01HG008383-01A1 to Y.K.] and supported by a National Science Foundation Award [DMS-1402254 to A.C.].

\section*{Appendix}
\appendix

\section{Experimental Details} \label{sec:experimental-details}
We run all linear CPH regression, Kaplan-Meier estimations, c-index statistics, and log-rank tests using the Lifelines Python package. DeepSurv is implemented in Theano with the Python package Lasagne. We use the R package randomForestSRC to fit RSFs. All experiments are run using Docker containers such that the experiments are easily reproducible. We use the FloydHub base image for the DeepSurv docker container.

The hyper-parameters of the network include: the depth and size of the network, learning rate, $\ell_2$ regularization coefficient, dropout rate, exponential learning rate decay constant , and momentum. We run the Random hyper-parameter optimization search as proposed in \citep{bergstra2012random} using the Python package Optunity. We use the Sobol solver \citep{sobol1976uniformly,Fox:1986:AIR:22721.356187} to sample each hyper-parameter from a predefined range and evaluate the performance of the configuration using $k$-means cross validation ($k=3$). We then choose the configuration with the largest validation C-index to avoid models that overfit. The hyper-parameters we use in all experiments are summarized in Appendix \ref{hyperparams}.

\subsection{Model Hyper-parameters} \label{hyperparams}

We tune DeepSurv's hyper-parameters by running a random hyper-parameter search using the Python package Optunity. The table below summarizes the hyper-parameters we use for each experiment's DeepSurv network. 

\def \linHPOptimizer {sgd}
\def \linHPActivation {SELU}
\def \linHPLayers {\num{1}}
\def \linHPNodes {\num{4}}
\def \linHPLR {\num{2.921891348405711}$\mathrm{e}{-4}$}
\def \linHPReg {\num{1.9989501953124997}}
\def \linHPDropout {\num{0.3752001953125}}
\def \linHPDecay {\num{3.57880859375}$\mathrm{e}{-4}$}
\def \linHPMomentum {\num{0.9064692382812499}}

\def \nonlinHPOptimizer {sgd}
\def \nonlinHPActivation {ReLU}
\def \nonlinHPLayers {\num{3}}
\def \nonlinHPNodes {\num{17}}
\def \nonlinHPLR {\num{3.1940075434571123}$\mathrm{e}{-4}$}
\def \nonlinHPReg {\num{4.4249755859375}}
\def \nonlinHPDropout {\num{0.4013037109375}}
\def \nonlinHPDecay {\num{3.17275390625}$\mathrm{e}{-4}$}
\def \nonlinHPMomentum {\num{0.9363432617187499}}

\def \whasHPOptimizer {adam}
\def \whasHPActivation {ReLU}
\def \whasHPLayers {\num{2}}
\def \whasHPNodes {\num{48}}
\def \whasHPLR {\num{0.0666363096295729}}
\def \whasHPReg {\num{16.094287109375}}
\def \whasHPDropout {\num{0.146552734375}}
\def \whasHPDecay {\num{6.494140624999997}$\mathrm{e}{-4}$}
\def \whasHPMomentum {\num{0.8631083984375}}

\def \supportHPOptimizer {adam}
\def \supportHPActivation{SELU}
\def \supportHPLayers {\num{1}}
\def \supportHPNodes {\num{44}}
\def \supportHPLR {\num{0.047277902766138385}}
\def \supportHPReg {\num{8.1196630859375}}
\def \supportHPDropout {\num{0.2553173828125}}
\def \supportHPDecay {\num{2.57333984375}$\mathrm{e}{-3}$}
\def \supportHPMomentum {\num{0.8589028320312501}}

\def \metabricHpOptimizer {adam}
\def \metabricHpActivation{SELU}
\def \metabricHpLayers {\num{1}}
\def \metabricHpNodes {\num{41}}
\def \metabricHpLR {\num{0.010289691253027908}}
\def \metabricHpReg {\num{10.890986328125}}
\def \metabricHpDropout {\num{0.160087890625}}
\def \metabricHpDecay {\num{4.1685546875}$\mathrm{e}{-3}$}
\def \metabricHpMomentum {\num{0.8439658203125}}

\def \trtHPOptimizer {adam}
\def \trtHPActivation{SELU}
\def \trtHPLayers {\num{1}}
\def \trtHPNodes {\num{45}}
\def \trtHPLR {\num{0.026024993217560365}}
\def \trtHPReg {\num{9.72212890625}}
\def \trtHPDropout {\num{0.10884765625000001}}
\def \trtHPDecay {\num{1.6355468750000002}$\mathrm{e}{-4}$}
\def \trtHPMomentum {\num{0.845416015625}}

\def \hormoneHpOptimizer {adam}
\def \hormoneHpActivation{SELU}
\def \hormoneHpLayers {\num{1}}
\def \hormoneHpNodes {\num{8}}
\def \hormoneHpLR {\num{0.153895727328729}}
\def \hormoneHpReg {\num{6.5512451171875}}
\def \hormoneHpDropout {\num{0.6606318359374999}}
\def \hormoneHpDecay {\num{5.667089843750001}$\mathrm{e}{-3}$}
\def \hormoneHpMomentum {\num{0.88674658203125}}

\begin{table}[h!]
\caption{DeepSurv's Experimental Hyper-parameters}
\label{table:hyperparams}
\begin{adjustbox}{center}
\begin{tabular}{ |c||c|c|c|c|c|c|c| }
 \hline
 Hyper-parameter & Sim Linear & Sim Nonlinear & WHAS & SUPPORT & METABRIC & Sim Treatment & GBSG \\
 \hline
 Optimizer & \linHPOptimizer & \nonlinHPOptimizer & \whasHPOptimizer & \supportHPOptimizer & \metabricHpOptimizer & \trtHPOptimizer & \hormoneHpOptimizer \\
 Activation & \linHPActivation & \nonlinHPActivation & \whasHPActivation & \supportHPActivation  & \metabricHpActivation & \trtHPActivation & \hormoneHpActivation\\
 \# Dense Layers & \linHPLayers & \nonlinHPLayers & \whasHPLayers & \supportHPLayers & \metabricHpLayers & \trtHPLayers & \hormoneHpLayers \\
 \# Nodes / Layer & \linHPNodes & \nonlinHPNodes & \whasHPNodes & \supportHPNodes & \metabricHpNodes &\trtHPNodes &\hormoneHpNodes  \\
 Learning Rate (LR) & \linHPLR & \nonlinHPLR & \whasHPLR & \supportHPLR & \metabricHpLR & \trtHPLR & \hormoneHpLR \\
 $\ell_2$ Reg & \linHPReg &\nonlinHPReg  & \whasHPReg & \supportHPReg & \metabricHpReg & \trtHPReg & \hormoneHpReg  \\
 Dropout & \linHPDropout & \nonlinHPDropout & \whasHPDropout &\supportHPDropout & \metabricHpDropout &\trtHPDropout & \hormoneHpDropout \\
 LR Decay & \linHPDecay & \nonlinHPDecay & \whasHPDecay & \supportHPDecay & \metabricHpDecay & \trtHPDecay& \hormoneHpDecay \\
 Momentum & \linHPMomentum & \nonlinHPMomentum & \whasHPMomentum & \supportHPMomentum & \metabricHpMomentum & \trtHPMomentum &\hormoneHpMomentum\\
 \hline
\end{tabular}
\end{adjustbox}
\end{table}

We applied inverse time decay to the learning rate at each epoch:
\begin{equation}
decayed\_LR := \frac{LR}{1 + epoch \cdot lr\_decay\_rate}.
\end{equation}

\section{CPH Recommender Function} \label{appendix:cph}
Let each patient in the dataset have a set of $n$ features $x_n$, in which one feature is a treatment variable $x_0 = \tau$. The CPH model estimates the risk function as a linear combination of the patient's features $\hat{h}_{\beta}(x) = \beta^Tx = \beta_0\tau + \beta_1 x_1 + ... + \beta_n x_n$. When we calculate the recommender function for the CPH model, we show that the model returns a constant function independent of the patient's features:
\begin{equation} \begin{aligned} \label{eq:cph_proof}
\rec_{ij}(x) &= \log \Big( \frac{\lambda(t;x | \tau = i)} {\lambda(t; x | \tau = j)} \Big) \\
&= \log \Big( \frac{\lambda_0(t) \cdot e^{\beta_0 i + \beta_1 x_1 + ... + \beta_n x_n}}{\lambda_0(t) \cdot e^{\beta_0 j + \beta_1 x_1 + ... + \beta_n x_n}} \Big) \\
&= \log \Big( e^{\beta_0 i + \beta_1 x_1 + ... + \beta_n x_n - (\beta_0 j + \beta_1 x_1 + ... + \beta_n x_n)} \Big) \\ 
&= \beta_0 i - \beta_0 j \\
&= \beta_0 (i-j).
\end{aligned}
\end{equation}
The CPH will recommend all patients to choose the same treatment option based on whether the model calculates the weight $\beta_0$ to be positive or negative. Thus, the CPH would not be providing personalized treatment recommendations. Instead, the CPH determines whether the treatment is effective and, if so, then recommending it to all patients. In an experiment, when we calculate which patients took the CPH's recommendation, the Recommendation and Anti-Recommendation subgroups will be equal to the control and treatment groups. Therefore, calculating treatment recommendations using the CPH provides little value to the experiments in terms of comparing the models' recommendations.


\section{Simulated Data Generation}\label{appendix:simulated}
Each patient's baseline information $x$ is drawn from a uniform distribution on $[-1,1)^d$. For datasets that also involve treatment, the patient's treatment status $\tau_x$ is drawn from a Bernoulli distribution with $p=0.5$.

The Cox proportional hazard model assumes that the baseline hazard function $\lambda_0(t)$ is shared across all patients. The initial death time is generated according to an exponential random variable with a mean $\mu=5$, which we denote $u\sim Exp(5)$. The individual death time is then generated by
\begin{eqnarray*}
T &=& \frac{u}{e^{h(x)}}, \textnormal{ when there is no treatment variable, }\\
T &=& \frac{u}{e^{\tau_x h(x)}}, \textnormal{ when there is a treatment variable}.
\end{eqnarray*}

These times are then right censored at an end time to represent the end of a trial. The end time $T_0$ is chosen such that 90 percent of people have an observed death time. 

Because we cannot observe any $T$ beyond the end time threshold, we denote the final observed outcome time
$$Z = \min(T, T_0).$$

\bibliographystyle{unsrt}
\bibliography{deep_survival}

\end{document}